\documentclass[a4paper,fleqn]{cas-sc}

\usepackage[numbers]{natbib}

\usepackage{amsthm}
\usepackage{float}
\usepackage[normalem]{ulem}
\usepackage{soul,color}
\usepackage{tablefootnote,gensymb}
\usepackage{eurosym}
\usepackage{tabularx}
\usepackage{svg}

\usepackage{algorithm}
\usepackage{algpseudocode}
\usepackage{tikz}
\usepackage{microtype}

\usepackage{caption}

\usepackage[utf8]{inputenc}
\usepackage{geometry}
\usepackage{titlesec}
\usepackage{tabu}                                                                                                     
\usepackage{longtable}
\usepackage{amsmath, amssymb}
\usepackage{xcolor}

\usepackage{tikz}
\usepackage{amsmath}
\usepackage{tikz-3dplot}
\usetikzlibrary{positioning}

\usepackage{hyperref}
\hypersetup{
    colorlinks=false,
    linkcolor=blue,
    filecolor=magenta,      
    urlcolor=blue,
}
\usepackage[binary-units=true]{siunitx}
\usepackage{dirtree}

\titleformat{\section}[block]{\hspace{1em}\bfseries}{\thesection.}{0.5em}{} 
\titleformat{\subsection}[block]{\hspace{1em}}{\thesubsection}{0.5em}{}
\newcolumntype{?}{!{\vrule width 2pt}}

\usepackage{subcaption}
\DeclareSymbolFont{Xlargesymbols}{OMX}{cmex}{m}{n}
\DeclareMathSymbol{\Xsum}{\mathop}{Xlargesymbols}{80}

\def\BibTeX{{\rm B\kern-.05em{\sc i\kern-.025em b}\kern-.08em
    T\kern-.1667em\lower.7ex\hbox{E}\kern-.125emX}}

\begin{document}
\let\WriteBookmarks\relax
\def\floatpagepagefraction{1}
\def\textpagefraction{.001}
\shorttitle{A Dataset of Images of Public Streetlights for Monitoring and Drift Detection}
\shortauthors{Li et~al.}

\title [mode = title]{A Multi-Year Urban Streetlight Imagery Dataset for Visual Monitoring and Spatio-Temporal Drift Detection}                      
\author[1]{Peizheng Li}[orcid=https://orcid.org/0000-0003-1516-1993]
\cormark[1]
\ead{peizheng.li@toshiba.eu}

\author[2]{Ioannis Mavromatis}[orcid=https://orcid.org/0000-0002-3309-132X]
\ead{ioannis.mavromatis@digicatapult.org.uk}

\author[1]{Ajith Sahadevan}[orcid=https://orcid.org/0009-0000-1046-7868]\ead{Ajith.Sahadevan@toshiba.eu}

\author[1]{Tim Farnham} [orcid=https://orcid.org/0000-0002-5355-3982]\ead{Tim.Farnham@toshiba.eu}

\author[1]{Adnan Aijaz}[orcid=https://orcid.org/0000-0003-1048-0469]\ead{adnan.aijaz@toshiba.eu}

\author[1]{Aftab Khan}[orcid=https://orcid.org/0000-0002-3573-6240]\ead{aftab.khan@toshiba.eu}

\address[1]{Bristol Research and Innovation Laboratory, Toshiba Europe Ltd., 30 Queen Square, Bristol, BS1 4ND, United Kingdom}
\address[2]{Digital Catapult, London NW1 2RA, UK}

\cortext[cor1]{Corresponding author}

\begin{abstract} 
We present a large-scale, longitudinal visual dataset of urban streetlights captured by 22 fixed-angle cameras deployed across Bristol, U.K., from 2021 to 2025. The dataset contains over 526,000 images, collected hourly under diverse lighting, weather, and seasonal conditions. Each image is accompanied by rich metadata, including timestamps, GPS coordinates, and device identifiers. This unique real-world dataset enables detailed investigation of visual drift, anomaly detection, and MLOps strategies in smart city deployments.
To promtoe seconardary analysis, we additionally provide a self-supervised framework based on convolutional variational autoencoders (CNN-VAEs). Models are trained separately for each camera node and for day/night image sets. We define two per-sample drift metrics: relative centroid drift, capturing latent space deviation from a baseline quarter, and relative reconstruction error, measuring normalized image-domain degradation. 
This dataset provides a realistic, fine-grained benchmark for evaluating long-term model stability, drift-aware learning, and deployment-ready vision systems. The images and structured metadata are publicly released in JPEG and CSV formats, supporting reproducibility and downstream applications such as streetlight monitoring, weather inference, and urban scene understanding. The dataset can be found at \url{https://doi.org/10.5281/zenodo.17781192} and \url{https://doi.org/10.5281/zenodo.17859120}.

\end{abstract}


\begin{keywords}

Streetlight \sep Computer Vision \sep Machine Learning \sep Smart Cities \sep Drift Detection \sep Variational Autoencoders \sep Ablation \sep Multi-year Dataset

\end{keywords}
\maketitle

\newpage\section*{Specifications table}
\vskip 0.2cm 
%
%
\renewcommand*{\arraystretch}{1.4}\begin{longtable}{|p{33mm}|p{124mm}|}
\hline
\textbf{Subject}                & Computer Science\\
\hline                         
\textbf{Specific subject area}  & Computer Vision, Drift Detection, Smart Cities Sensing \\

\hline
\textbf{Type of data}           & Raw data in JPEG format and structured tabular data in CSV format. Each zipped folder contains image samples from a single lamppost camera (serial ID). The main CSV file includes metadata (timestamp, GPS, image name), photometric statistics, day/night status, and two drift metrics—relative centroid drift and relative reconstruction error. All metadata fields are provided in plain text, suitable for visual inspection, scripting, and downstream ML pipelines. Images are collected from 22 nodes and recorded over four years.

\\                                   
\hline
\textbf{Data collection} & Data were collected from 22 UMBRELLA nodes, each mounted on public streetlights in Bristol, UK. Each node includes a Raspberry Pi with Camera Module V1 (OV5647 sensor), capturing upward-facing images of the light column and sky. Images were collected hourly from 2021 to 2025 via the UMBRELLA testbed~\cite{umbrella_paper,UMBRELLA}. Images were captured approximately once per hour, 24/7, for a period of four years (2021–2025). A randomised capture delay (up to 600 seconds) was added to reduce server load. All images were uploaded to a central server for indexing and post-processing.



\\                         
\hline                         
\textbf{Data source location}   & Around 7.2 km of public road in Bristol and South Gloucestershire, UK, including Coldharbour Lane, Filton Road, and the A4174 Ring Road. Additional nodes are located at the University of the West of England (Frenchay Campus).

\\
\hline                         
\hypertarget{target1}
{\textbf{Data accessibility}}   & Data available online at the links below: \newline \url{https://doi.org/10.5281/zenodo.17781192}~\cite{li_dataset_part1}
\newline \url{https://doi.org/10.5281/zenodo.17859120}~\cite{li_dataset_part2}

\\                         
\hline                         
\textbf{Related                 
research\newline
article}                &   None
\\
\hline                         
\end{longtable}

\section{Value of the Data}

This work presents a large-scale, long-term visual dataset of streetlight imagery, purpose-built for research in urban infrastructure monitoring and Machine Learning-based drift detection. Its distinctive attributes make it valuable across a wide range of Computer Vision and Smart Cities applications. The key contributions are as follows:

\begin{itemize}
    \item \textbf{Large-Scale, Longitudinal Visual Dataset:}  
    The dataset comprises \textbf{526,096 JPEG images} collected between \textbf{2021 and 2025} from the city of Bristol, UK. Each image is captured by one of 22 fixed-angle Raspberry Pi Camera modules mounted on public lampposts. These devices provide diverse viewpoints with varying occlusions, scene compositions, and lighting conditions. The extended temporal span enables the analysis of both natural and operational distribution shifts, supporting tasks such as \emph{drift detection}, \emph{anomaly detection}, and long-term model evaluation.

    \item \textbf{Diverse Detection Scenarios:}  
    Streetlights (and the nodes mounted on them) are manually classified into \textbf{Type-0 (ideal)} and \textbf{Type-1 (challenging)} categories, based on lamp visibility within the Field of View (FoV) of the camera. Type-0 nodes offer clear lamp visibility, enabling straightforward detection, whereas Type-1 nodes involve occlusions, misalignments, or poor lighting conditions. This distinction allows for robust benchmarking across varying degrees of difficulty.

    \item \textbf{Drift Detection Benchmark with Real-World Deployment:}  
    We introduce a drift detection benchmark based on natural variations in urban scenes. The methodology includes: (1) quarterly temporal segmentation; (2) CNN-VAE training per day/night split and camera node; and (3) latent space evaluation using centroid drift and reconstruction error. An ablation across multiple latent dimensions quantifies sensitivity to representation capacity, enabling fair and reproducible comparisons.

    \item \textbf{Relevance to MLOps and Continual Learning:}  
    The dataset reflects real-world visual evolution due to weather, lighting, and seasonal changes, providing a valuable testbed for validating MLOps strategies such as drift-triggered retraining, degradation monitoring, and lightweight continual learning models for edge devices. A number of potential use cases are described in~\cite{li2024past}.

    \item \textbf{Multimodal and Practical Applications:}  
    Beyond drift detection, this dataset supports diverse Smart Cities applications, including:
    \begin{itemize}
        \item Streetlight status recognition, classifying ON/OFF states for energy and maintenance optimisation;
        \item Weather condition inference, e.g., fog or rain detection from image quality and noise;
        \item Transfer learning for urban perception, due to the rich variations in lighting, occlusion, and environment.
    \end{itemize}

    \item \textbf{Utility for Representation Learning:}  
    The dataset’s variability in illumination, weather, and viewing angle makes it suitable for unsupervised representation learning. Pretraining on this data can yield robust, generalisable features for downstream Smart Cities vision tasks.
\end{itemize}

\section{Background}
\begin{figure*}   
    \subfloat[\label{fig:node0}]{
      \begin{minipage}[t]{0.5\linewidth}
        \centering 
        \includegraphics[width=3.2in]{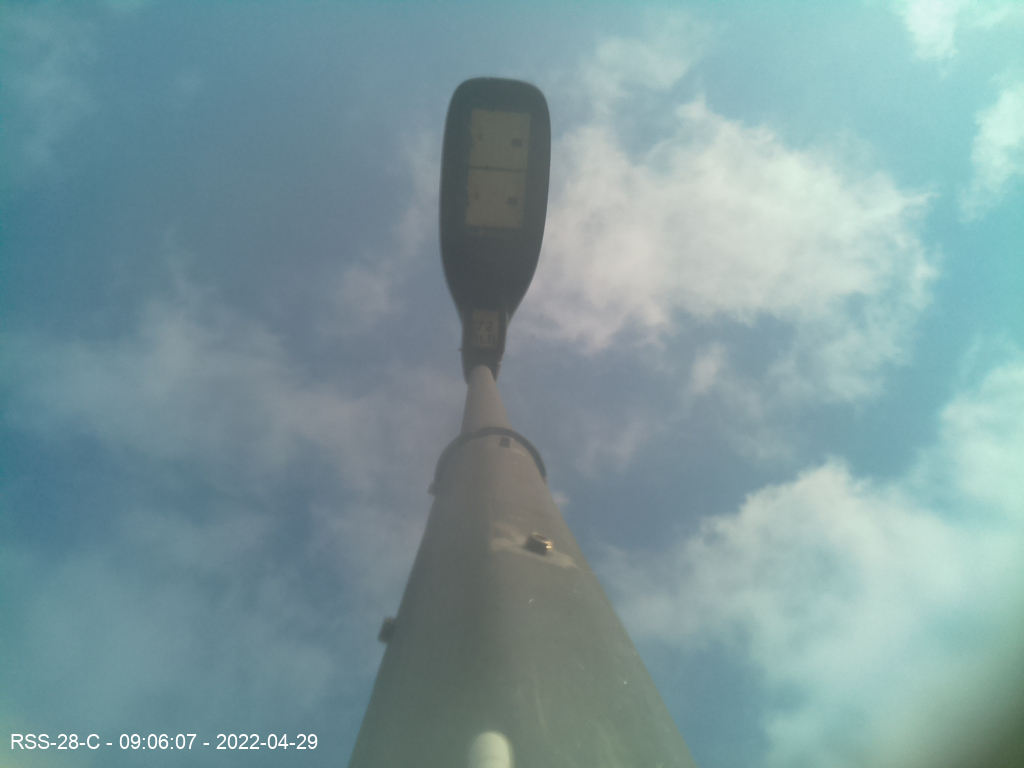}   
      \end{minipage}%
      }
        \subfloat[\label{fig:node1}]{
      \begin{minipage}[t]{0.5\linewidth}   
        \centering   
        \includegraphics[width=3.2in]{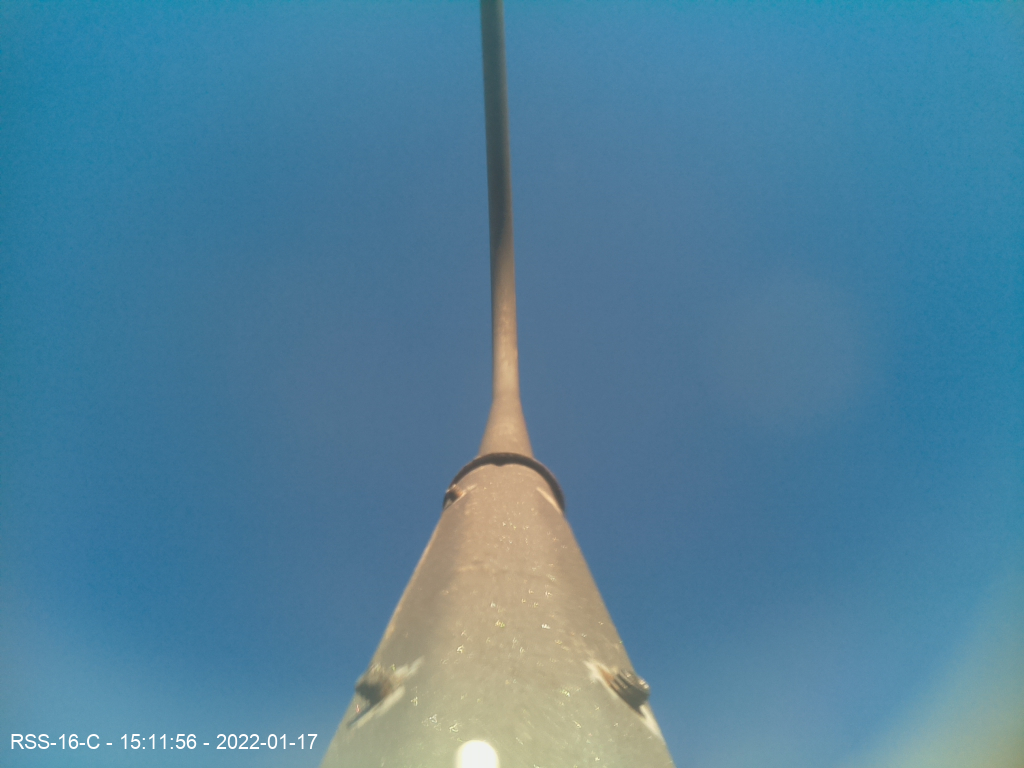}   
      \end{minipage} 
      }
      \caption{(a) Type 0: Ideal lamppost in which the light is clearly visible; (b) Type 1: In this type, the light is not directly visible.
      } \label{fig:2nodes}
      \vspace{-0.2cm}
\end{figure*} 
The primary objective of this work is to introduce and open-source a large-scale, diverse, and temporally rich visual dataset of urban streetlights, captured via fixed-angle cameras installed on public lighting infrastructure across Bristol, UK. Spanning over \textbf{526,000 images} collected continuously between \textbf{2021 and 2025}, the dataset encompasses a wide spectrum of lighting conditions, seasonal changes, and weather-induced variations.

This dataset is designed to serve as a realistic and challenging benchmark for visual inspection, infrastructure monitoring, and maintenance research within Internet-of-Things (IoT) environments. Its uniqueness lies in its diverse FoV constraints and environmental artefacts such as rain, fog, and glare, which introduce complex edge cases ideal for advancing anomaly detection and robust classification under real-world deployment conditions.

To demonstrate its utility, we propose a drift detection framework based on Convolutional Variational Autoencoders (CNN-VAEs), leveraging two key metrics: latent space \emph{centroid drift} and \emph{relative reconstruction error}. These metrics are computed on a quarterly basis and provide fine-grained insight into evolving data distributions. Ablation studies across different latent dimensionalities and camera node types (e.g., visible vs. occluded lamp views) further explore the sensitivity of learned representations to environmental and architectural factors.

Overall, this dataset provides a valuable foundation for research in visual drift detection, unsupervised representation learning, model retraining policy development, and MLOps methodologies targeting long-term, real-world smart city deployments.

\section{Data Description}

The dataset is published across two Zenodo records~\cite{li_dataset_part1,li_dataset_part2}.
Record~\cite{li_dataset_part1} contains node-specific archives for 18 streetlight nodes together with the central metadata file (streetcare-drift-dataset-2021-2025.csv), whereas record~\cite{li_dataset_part2} includes the archives for the remaining four nodes. Each archive is a standalone .zip file that can be extracted and used independently.

Upon extraction, each archive yields a collection of JPEG images whose filenames are unique 32-character alphanumeric strings (e.g., f2a4b…c.jpg). These identifiers are randomly generated at capture time and remain stable across all processing stages. An example directory structure of the dataset is illustrated below:

\vskip 0.5cm

\noindent\makebox[\textwidth][c]{%
    \begin{varwidth}{0.5\textwidth}
        \dirtree{%
          .1 streetcare-drift-dataset-2021-2025 \cite{li_dataset_part1,li_dataset_part2} .
          .2 streetcare-drift-dataset-2021-2025.csv.
          .2 RSE-19-C.zip.
          .3 0a0e7888ad51010cffdc14f20ede9564.jpg.
          .3 00a0f83fda98ab9b586bee3a9335e4dd.jpg.
          .3 ....
          .2 RSE-34-C.zip.
          .3 0a0dbd9aace69a306bf73d36ed5daa4d.jpg.
          .3 0a1c9bc590590e7536d43c4a5a898e95.jpg.
          .3 ....
          .2 ....
        }
    \end{varwidth}
}

\vskip 0.5cm

All images were captured using the same Raspberry Pi Camera Module V1~\cite{picamera}, installed across 22 UMBRELLA nodes (i.e., lamppost-mounted devices). Each node offers a distinct viewpoint, due to variations in placement, height, orientation, and environmental occlusion. The dataset comprises a total of \textbf{526,096 images}, with \textbf{270,267 daytime} and \textbf{255,829 nighttime} captures, spanning a period of five years from 2021 to 2025. Images are further categorised into two visibility types (as shown in Fig.~\ref{fig:2nodes}): Type-0 (\textbf{ideal nodes}, 258,527 images) and Type-1 (\textbf{challenging nodes}, 267,569 images), each encompassing data from 11 serial IDs, as summarised in~Table~\ref{tab:dataset}.

Each UMBRELLA node captures an image approximately once per hour, with a randomised delay of up to 600 seconds to balance backend load. Captured images are transmitted to a backend server and indexed in a structured CSV file (each row in the CSV file represents an ``entry'' in the dataset), which includes metadata fields such as timestamp, serial ID, GPS coordinates, hostname, lighting condition (day/night), and fault status.

The CSV file also includes two important drift-related fields: \texttt{relative\_centroid\_drift} and \texttt{relative\_recon\_error}. These metrics enable per-sample drift analysis and are computed using a CNN-based Variational Autoencoder (CNN-VAE). Each image is resized to $64 \times 64 \times 3$ and passed through a trained CNN-VAE with a latent dimension of $d = 16$. The latent vectors $\mathbf{z}_i$ are grouped quarterly by timestamp (\texttt{time\_tag}) to track temporal drift. A baseline centroid is computed from the first quarter's latent vectors. 

\begin{itemize}
    \item \textbf{Relative centroid drift} is calculated as the normalised Euclidean distance between each image’s latent vector and the baseline centroid.
    \item \textbf{Relative reconstruction error} measures the deviation of each image's reconstruction loss from the baseline quarter’s average.
\end{itemize}

Both metrics allow granular, longitudinal quantification of distributional shift in both latent and image domains. They are particularly useful for MLOps-related applications~\cite{li2024adapting} such as drift monitoring, retraining triggers, and continual learning policy evaluation. Users can recompute drift metrics using their own grouping strategy (e.g., monthly, biannual), since image-to-metadata alignment is maintained via the \texttt{image\_name} field.

\begin{figure}[t]
    \centering
    \includegraphics[width=1\textwidth]{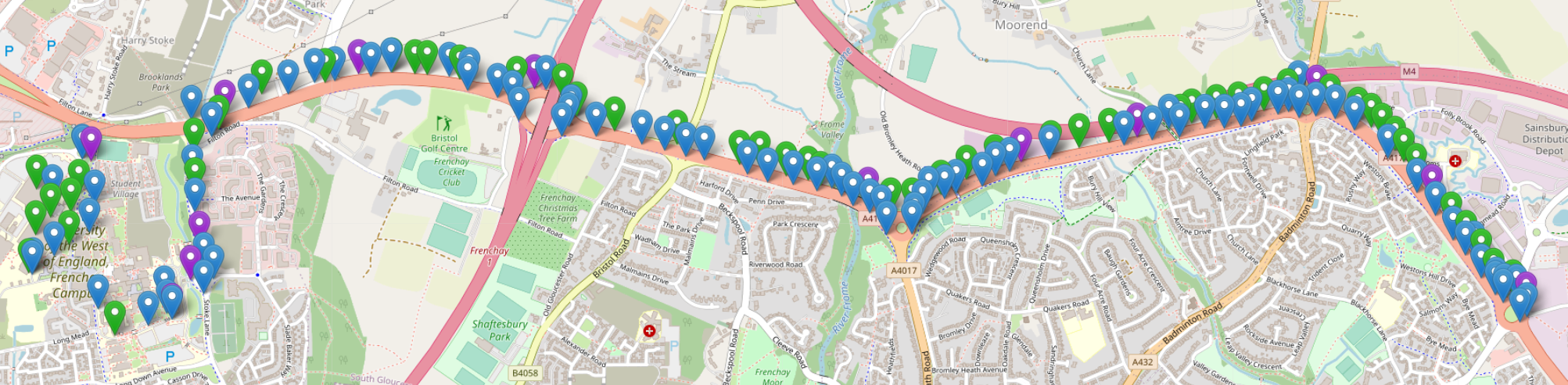}
    \caption{The UMBRELLA network. Nodes are installed on public lampposts along a $\sim$\SI{7.2}{\kilo\meter} road. Colours indicate connectivity: green = fibre, blue = WiFi, purple = fibre + LoRa gateway.}
    \label{fig:umbrella_network}
\end{figure}

\begin{figure*}   
    \subfloat[\label{fig:lamp1}]{
      \begin{minipage}[t]{0.5\linewidth}
        \centering 
        \includegraphics[width=3.2in]{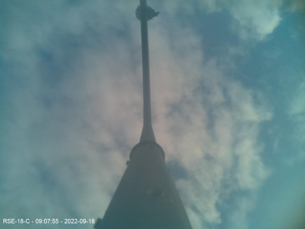}   
      \end{minipage}%
      }
        \subfloat[\label{fig:lamp2}]{
      \begin{minipage}[t]{0.5\linewidth}   
        \centering   
        \includegraphics[width=3.2in]{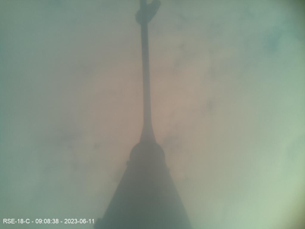}   
      \end{minipage} 
      }
      \\
    \subfloat[\label{fig:lamp3}]{
      \begin{minipage}[t]{0.5\linewidth}
        \centering 
        \includegraphics[width=3.2in]{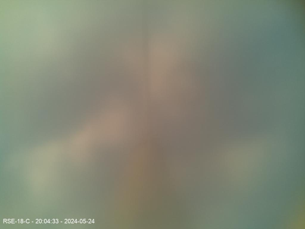}   
      \end{minipage}%
      }
        \subfloat[\label{fig:lamp4}]{
      \begin{minipage}[t]{0.5\linewidth}   
        \centering   
        \includegraphics[width=3.2in]{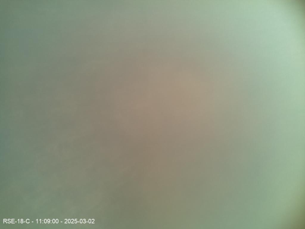}   
      \end{minipage} 
      }

      \caption{Example images captured by RSE-18-C in 2022 (a), 2023 (b), 2024 (c), 2025 (d). 
      } 
      \label{fig:4_lampes}
\end{figure*}

All data were collected through the UMBRELLA testbed described in~\cite{umbrella_paper}, whose deployment topology is visualised in Fig.~\ref{fig:umbrella_network}. The nodes are deployed along a \SI{7.2}{\kilo\meter} urban corridor, covering diverse street and lighting conditions. A representative comparison of the same lamppost over four years is illustrated in Fig.~\ref{fig:4_lampes}, showing natural degradation and increasing blur due to long-term wear, environmental exposure, and optical drift.

\begin{table}[htbp]
\centering
\caption{Dataset IDs, sizes, and number of images by the type of node}
\label{tab:dataset}
\begin{tabular}{|c|c|c|c|c?c|c|c|c|c|}
\hline
\textbf{Type}                & \textbf{Serial ID} & \textbf{Size (GB)} & \textbf{DayImgs} & \textbf{NightImgs} & \textbf{Type}                & \textbf{Serial ID} & \textbf{Size (GB)} & \textbf{DayImgs} & \textbf{NightImgs} \\ \hline
\multirow{11}{*}{\textbf{0}} & RSE-19-C           & 12                 & 13,743                      & 12,502                        & \multirow{11}{*}{\textbf{1}} & RSE-34-C           & 11                 & 12,723                      & 11,688                        \\ \cline{2-5} \cline{7-10} 
                             & RSS-A-2-C          & 9.9                & 11,026                      & 10,858                        &                              & RSE-44-C           & 12                 & 13,361                      & 12,695                        \\ \cline{2-5} \cline{7-10} 
                             & RSE-22-C           & 9.8                & 11,230                      & 10,604                        &                              & RSE-A-11-C         & 12                 & 12,970                      & 12,027                        \\ \cline{2-5} \cline{7-10} 
                             & RSE-21-C           & 9.7                & 12,133                      & 11,501                        &                              & RSE-A-15-C         & 12                 & 13,550                      & 12,379                        \\ \cline{2-5} \cline{7-10} 
                             & RSE-25-C           & 9.7                & 11,087                      & 10,360                        &                              & RSS-15-C           & 11                 & 11,981                      & 11,547                        \\ \cline{2-5} \cline{7-10} 
                             & RSE-24-C           & 9.0                & 12,631                      & 11,905                        &                              & RSS-47-C           & 11                 & 11,722                      & 11,340                        \\ \cline{2-5} \cline{7-10} 
                             & RSS-12-C           & 8.7                & 11,788                      & 11,868                        &                              & RSS-58-C           & 12                 & 12,862                      & 12,153                        \\ \cline{2-5} \cline{7-10} 
                             & RSE-6-C            & 8.3                & 12,968                      & 11,707                        &                              & RSS-59-C           & 11                 & 11,956                      & 11,552                        \\ \cline{2-5} \cline{7-10} 
                             & RSE-16-C           & 8.1                & 10,269                      & 9,855                         &                              & RSS-8-C            & 11                 & 11,647                      & 11,245                        \\ \cline{2-5} \cline{7-10} 
                             & RSE-A-10-C         & 8.1                & 12,328                      & 12,124                        &                              & RSS-A-13-C         & 11                 & 12,094                      & 11,599                        \\ \cline{2-5} \cline{7-10} 
                             & RSE-A-7-C          & 7.7                & 13,677                      & 12,363                        &                              & RSS-A-33-C         & 11                 & 12,521                      & 11,957                        \\ \hline
\end{tabular}
\end{table}

\begin{table}[t]
\centering
\caption{Definition of the labels of each column of the CSV data files in the dataset.}
\label{tab:csvlabels}
\renewcommand{\arraystretch}{1.2}
\begin{tabularx}{\textwidth}{r|>{\raggedright\arraybackslash}X}
\hline
\textbf{Field Name} & \textbf{Definition} \\
\hline \hline
\texttt{id} & A unique ID given to each node. It increments by one for each image captured. \\
\texttt{serial} & The UMBRELLA node's ``serial ID'' (friendly name). \\
\texttt{date} & Date and timestamp at the moment an image was captured. \\
\texttt{hostname} & Hostname of the UMBRELLA node as seen in the OS. \\
\texttt{lat} & Latitude of the node location. \\
\texttt{lon} & Longitude of the node location. \\
\texttt{image\_name} & JPEG image file name (32-digit alphanumeric string ending in `.jpg`). \\
\texttt{time\_tag} & Quarter when the image was captured (e.g., 2023-Q1). \\
\texttt{fault\_detected} & The post-processed light status (operational or not). \\
\texttt{confidence} & The confidence score for the fault detection result. \\
\texttt{daynight} & Indicates whether the image was captured during the day or night cycle. \\
\texttt{red} & Mean pixel intensity in the red channel. \\
\texttt{green} & Mean pixel intensity in the green channel. \\
\texttt{blue} & Mean pixel intensity in the blue channel. \\
\texttt{relative\_centroid\_drift} & Normalized distance between a sample’s latent vector and the baseline quarter’s centroid. \\
\texttt{relative\_recon\_error} & Normalized reconstruction error relative to the baseline quarter’s average. \\
\hline
\end{tabularx}
\end{table}

\section{Drift Detection Experimental Methods and Results}

\begin{figure}[t]
    \centering
    \includegraphics[width=1\textwidth]{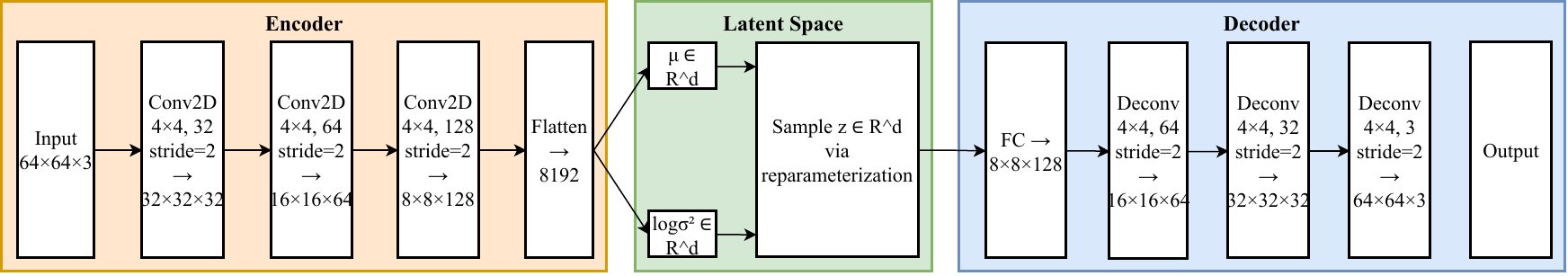}
    \caption{VAE architecture for drift detection.}
    \label{fig:vae_architecture}
\end{figure}


\begin{figure}[t]
    \centering
    \includegraphics[width=1\textwidth]{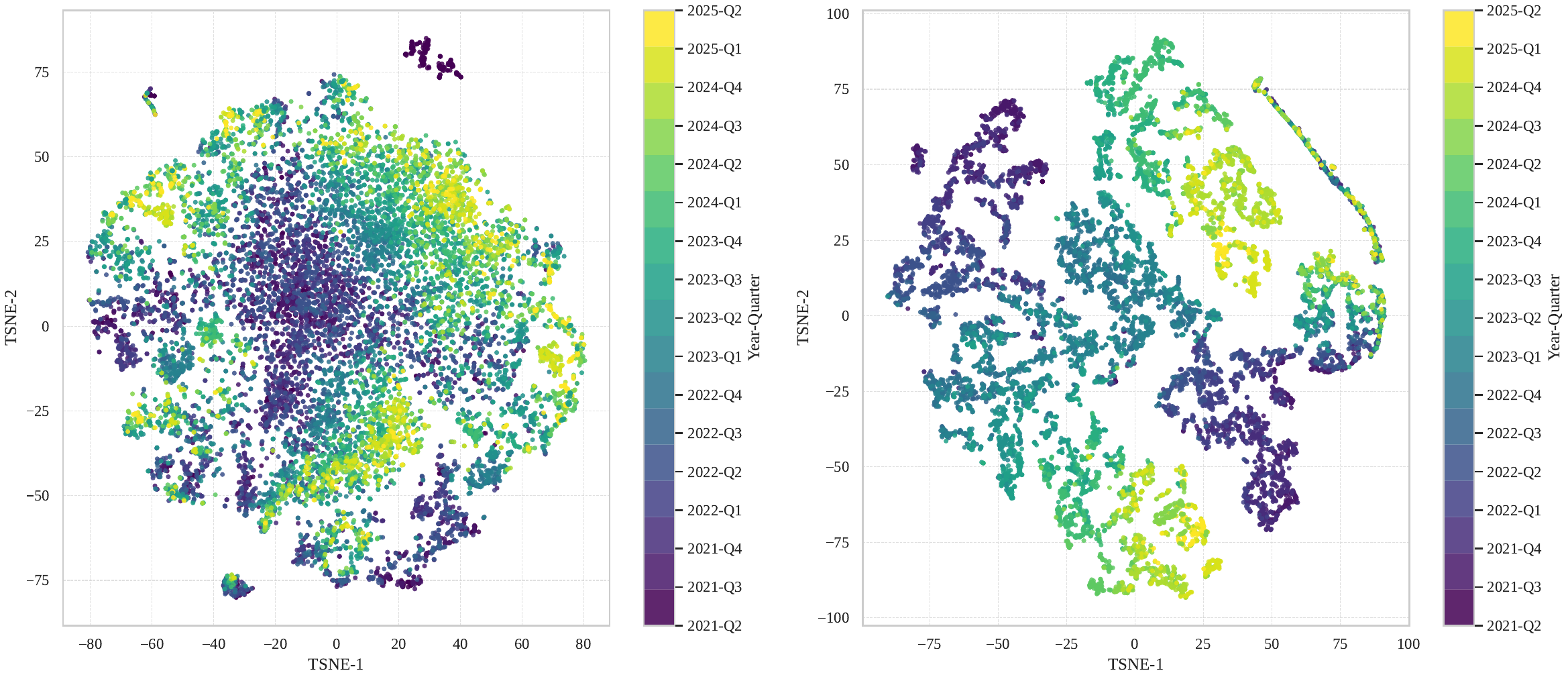}   
      \caption{Visualisation of the latent space distribution for the day dataset (left) and the night dataset (right) of serial ID RSE-19-C. 
      } \label{fig:2figures_latent_space}
\end{figure}


\subsection{Analysis Methods, Models, and Drift Metrics}
\label{subsec:model details}

To illustrate the utility of our dataset, we conduct a detailed camera drift analysis leveraging its long-term, high-frequency visual data. Given the extended temporal span and uncontrolled outdoor conditions, the dataset inherently exhibits non-stationary visual distributions. For our analysis, we segment the data by UMBRELLA node (using serial ID) and by lighting condition (day/night), with each image annotated by its timestamp and metadata.

We adopt a self-supervised drift detection framework based on a CNN-VAE. A separate CNN-VAE model is trained for each node and lighting condition pair. All images are resized to $64 \times 64 \times 3$ before training. The encoder compresses input images into a latent vector $\mathbf{z} \in \mathbb{R}^d$, where $d$ is the latent dimensionality. We conduct ablation over $d \in \{4, 8, 16, 32\}$. The decoder reconstructs the original image, and the model is trained using the standard VAE loss function: a combination of pixel-wise reconstruction error and KL divergence regularization. The model architecture is illustrated in Fig.~\ref{fig:vae_architecture}.

To quantify drift, we compute latent and reconstruction-based metrics over quarterly temporal groups. As shown in Fig.~\ref{fig:2figures_latent_space}, the latent space distributions for RSS-A-38-C (with $d = 32$) show clear separation between different quarters for both day and night images, suggesting a significant temporal shift. We define two complementary metrics:

\begin{itemize}
    \item \textbf{Centroid Drift (CD):} For each quarter, we compute the mean latent vector (centroid) across samples. Drift is defined as the Euclidean distance between a given quarter's centroid and the baseline centroid (from the first quarter).
    \item \textbf{Relative Reconstruction Error (RE):} For each quarter, we compute the mean pixel-wise reconstruction error and define drift as the relative increase from the baseline (first quarter's) average.
\end{itemize}

Both metrics are globally normalized to enable comparisons across nodes, lighting conditions, and model configurations. Boxplots are used to visualize the distribution of these drift values across all 22 nodes, with separate analyses by latent dimension, day/night split, and visibility type (Type-0 vs. Type-1).

\subsection{Results and Analysis}
\begin{figure}[t]
    \centering
    \includegraphics[width=1\textwidth]{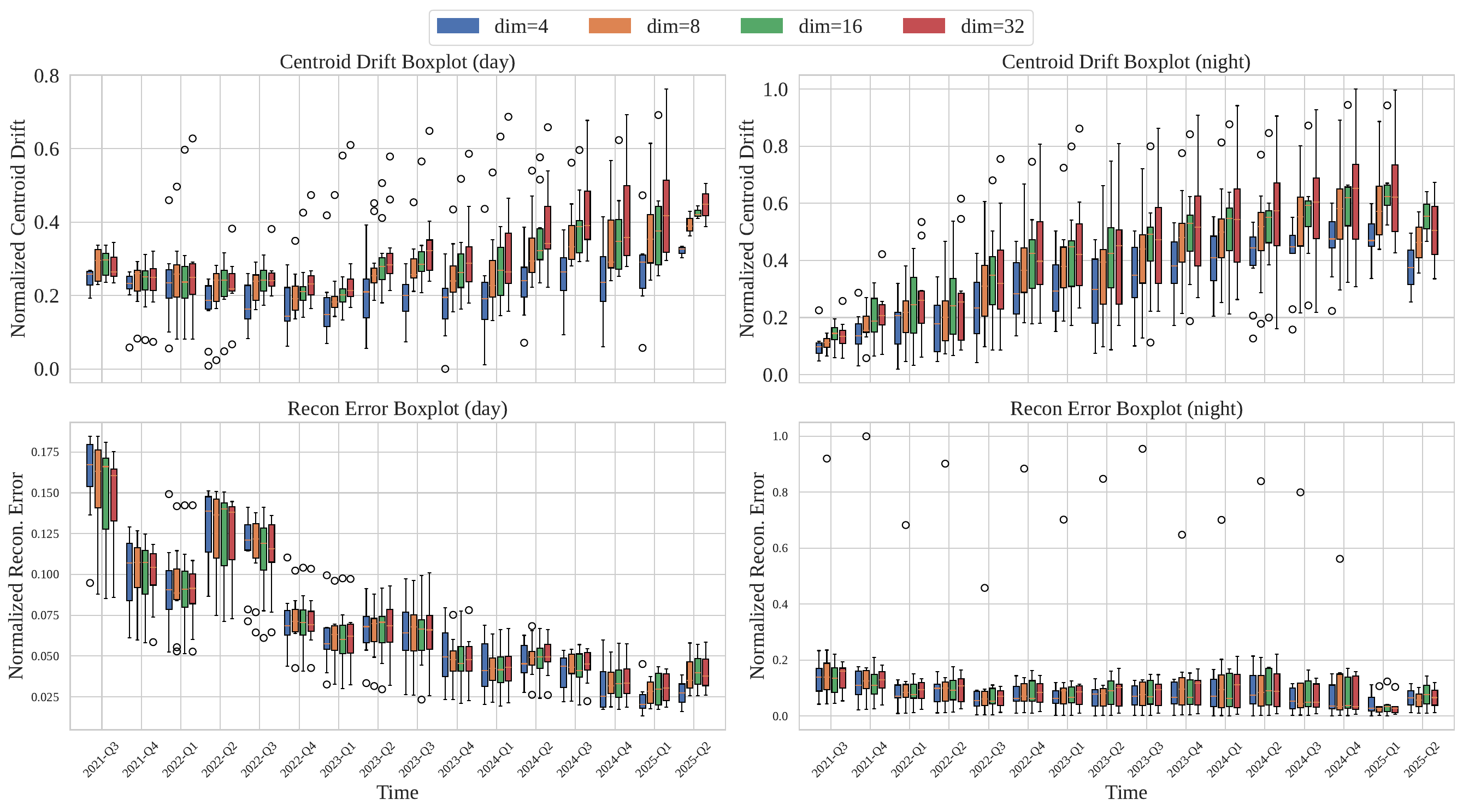}
    \caption{Boxplots of centroid drift and reconstruction error for ablation study results of type 0 serials.}
    \label{fig:type0_boxplot}
\end{figure}

\begin{figure}[t]
    \centering
    \includegraphics[width=1\textwidth]{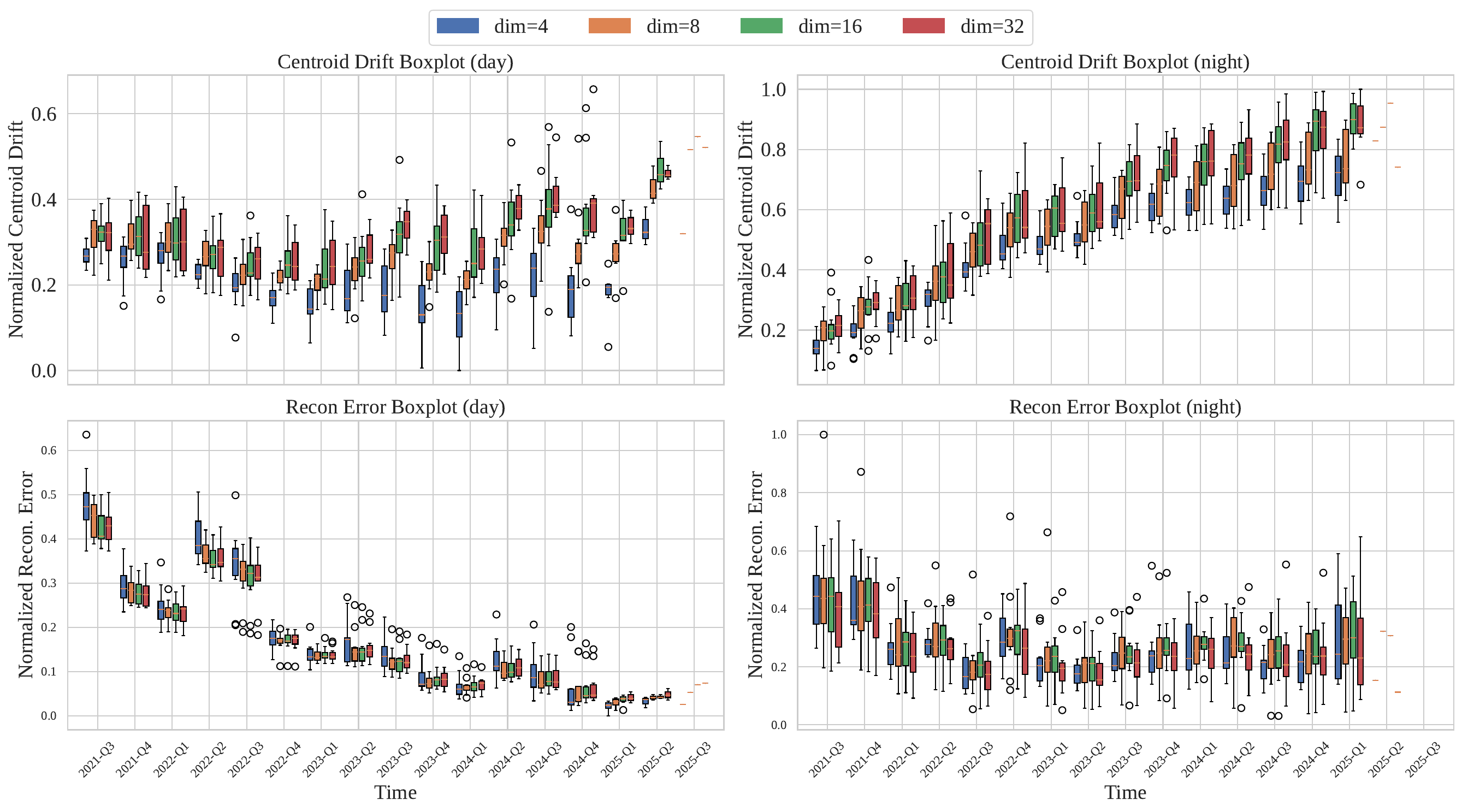}
    \caption{Boxplots of centroid drift and reconstruction error for ablation study results of type 1 serials.}
    \label{fig:type1_boxplot}
\end{figure}
We trained a total of 44 CNN-VAE models, one for each of the 22 camera nodes, further split by day and night conditions. Each model was trained independently using 200 epochs, an Adam optimizer with a learning rate of $10^{-3}$, and a batch size of 1024.

Post-training, we extracted latent vectors and reconstruction errors for all images. The data was temporally grouped into quarterly segments (i.e., 3-month intervals), yielding approximately 16 time bins for drift analysis. Results are visualized in Fig.~\ref{fig:type0_boxplot} and Fig.~\ref{fig:type1_boxplot}, using boxplots to depict the distribution of normalized drift metrics across quarters. Each plot overlays results across four latent dimensions ($d \in \{4, 8, 16, 32\}$), separately for day/night splits and drift metrics.

\begin{itemize}
    \item \textbf{Type-0 Nodes (Ideal FoV):}
    \begin{itemize}
        \item Centroid drift remains consistently low, particularly for smaller latent dimensions and daytime images, suggesting that low-dimensional latent spaces are sufficient for modeling stable visual semantics under favorable conditions.
        \item Larger latent dimensions (e.g., $d=32$) introduce higher variance over time, possibly due to overfitting to minor visual changes or noise.
        \item Reconstruction error exhibits a downward trend over time, indicating either lighting stabilization or reduced variability in captured scenes.
    \end{itemize}
    
    \item \textbf{Type-1 Nodes (Challenging FoV):}
    \begin{itemize}
        \item Centroid drift increases steadily across quarters, particularly under night conditions. This trend likely reflects cumulative effects from occlusion, environmental change, or sensor aging.
        \item The increase in drift is more pronounced at higher latent dimensions, suggesting increased sensitivity to subtle or noisy variations.
        \item Reconstruction error varies non-monotonically, especially at night, likely due to inconsistent visibility or scene complexity.
    \end{itemize}
\end{itemize}

These findings underscore the trade-off between model expressiveness and robustness. While larger latent spaces can capture fine-grained changes, they are also more susceptible to overfitting and environmental noise. For long-term deployment scenarios such as Smart Cities monitoring or MLOps-driven retraining, a moderate latent dimension ($d=8$ or $d=16$) may offer the best balance between sensitivity and stability.

Overall, the results validate this dataset as a valuable benchmark for studying visual drift in real-world, long-term deployments. The proposed CNN-VAE framework, coupled with interpretable drift metrics, provides a scalable foundation for longitudinal model evaluation under realistic conditions.

\section{Limitations}
The current release covers 22 upward-facing camera nodes deployed across a single UK municipality with identical hardware, mounting heights, and firmware. The annotations that accompany the images are intentionally lightweight: apart from node metadata, illumination regime, and automated quality-control tags, we do not provide exhaustive manual labels of lamp faults or seasonal events. This constrains the dataset to predominantly unsupervised or weakly supervised use cases, and practitioners seeking strongly labeled benchmarks for downstream detection or classification tasks will need to create their own annotations or fuse the imagery with municipal work-order records.

\section*{Ethics statements}
Hereby, the authors consciously assure that for the manuscript “A Multi-Year Urban Streetlight Imagery Dataset for Visual Monitoring and Spatio-Temporal Drift Detection” the following is fulfilled:
\begin{enumerate}
    \item This material is the authors' own original work, which has not been previously published elsewhere.
    \item The paper is not currently being considered for publication elsewhere.
    \item The paper reflects the authors' own research and analysis in a truthful and complete manner.
    \item The paper properly credits the meaningful contributions of co-authors and co-researchers.
    \item The results are appropriately placed in the context of prior and existing research.
    \item All sources used are properly disclosed (correct citation). Literally copying of text must be indicated as such by using quotation marks and giving proper reference.
    \item All authors have been personally and actively involved in substantial work leading to the paper, and will take public responsibility for its content.
\end{enumerate}

As our dataset did not involve any human subjects, animal experiments, or social media platform data, approval from any IRB/local ethics committees was not required. As our camera images are facing the sky, no human subjects are present in the photos. Finally, as our dataset is based on street light images, no survey studies were conducted, and no work was conducted involving chemicals, procedures, or equipment that have any usual hazards inherent in their use, against aminal or human subjects.

I agree with the above statements and declare that this submission follows the policies of Solid State Ionics as outlined in the Guide for Authors and in the Ethical Statement.


\section*{CRediT author statement}
\textbf{Peizheng Li}: Conceptualization, Methodology, Software, Data Curation, Writing - Original Draft, 
\textbf{Ioannis Mavromatis}: Conceptualization, Software, Data Curation, Writing - Review,
\textbf{Ajith Sahadevan}: Software, Data Curation, Writing - Review, 
\textbf{Tim Farnham}: Methodology, Writing - Review, 
\textbf{Adnan Aijaz}: Methodology, Writing - Review, 
\textbf{Aftab Khan}: Conceptualization, Methodology, Writing - Review \& Editing.


\section*{Acknowledgments}
This work is funded in part by Toshiba Europe Ltd. UMBRELLA project is funded in conjunction with South Gloucestershire Council by the West of England Local Enterprise Partnership through the Local Growth Fund, administered by the West of England Combined Authority.
\newline

\section*{Declaration of Competing Interest}

\begin{itemize}
\item[$\checkmark$]{The authors declare that they have no known competing financial interests or personal relationships that could have appeared to influence the work reported in this paper.}

\item[$\square$]{The authors declare the following financial interests/personal relationships which may be considered as potential competing interests: }
\end{itemize}

\bibliographystyle{IEEEtran}
\bibliography{refs}

@misc{UMBRELLA,
  author={{BRIL Toshiba Europe Ltd.}},
  title = {{UMBRELLA Platform and Testbed}},
  year={2022},
  howpublished = {\url{https://www.umbrellaiot.com}},
  note = {{Accessed: 2022-1-31}}
}

@article{umbrella_paper,
  title = {{UMBRELLA: A One-stop Shop Bridging the Gap from Lab to Real-World IoT Experimentation}},
  author = {{Mavromatis}, I. and {Jin}, Y. and {Stanoev}, A. and {Portelli}, A. and {Weeks}, I. and {Holden}, B. and {Glasspole}, E. and {Farnham}, T. and {Khan}, A. and {Raza}, U. and {Aijaz}, A. and {Bierton}, T. and {Seto}, I. and {Patel}, N. and {Sooriyabandara}, M.},
  journal = {IEEE Access},
  year = {2024},
  month = mar,
  doi = {10.1109/ACCESS.2024.3377662}
}

@misc{picamera,
  author={{Raspberry Pi}},
  year={2015},
  title = {{Raspberry Pi Camera Module ver.1 Datasheet}},
  howpublished = {\url{https://www.arducam.com/downloads/modules/RaspberryPi_camera/RaspberryPi_Camera_Module_DS_V3.0.pdf}},
  note = {{Accessed: 2022-1-31}}
}

@inproceedings{li2024past,
  title={Past, present, future: A comprehensive exploration of ai use cases in the umbrella iot testbed},
  author={Li, Peizheng and Mavromatis, Ioannis and Khan, Aftab},
  booktitle={2024 IEEE International Conference on Pervasive Computing and Communications Workshops and other Affiliated Events (PerCom Workshops)},
  pages={787--792},
  year={2024},
  organization={IEEE}
}

@article{li2024adapting,
  title={Adapting MLOps for Diverse In-Network Intelligence in 6G Era: Challenges and Solutions},
  author={Li, Peizheng and Mavromatis, Ioannis and Farnham, Tim and Aijaz, Adnan and Khan, Aftab},
  journal={arXiv preprint arXiv:2410.18793},
  year={2024}
}

@dataset{li_dataset_part1,
  author       = {Li, Peizheng and
                  Mavromatis, Ioannis and
                  Ajith, Sahadevan and
                  Farnham, Tim and
                  Aijaz, Adnan and
                  Khan, Aftab},
  title        = {Multi-Year Urban Streetlight Image Collection for
                   Visual Monitoring and Drift Analysis (Part 1/2)
                  },
  month        = dec,
  year         = 2025,
  publisher    = {Zenodo},
  doi          = {10.5281/zenodo.17781192},
  url          = {https://doi.org/10.5281/zenodo.17781192},
}

@dataset{li_dataset_part2,
  author       = {Li, Peizheng and
                  Mavromatis, Ioannis and
                  Ajith, Sahadevan and
                  Farnham, Tim and
                  Aijaz, Adnan and
                  Khan, Aftab},
  title        = {Multi-Year Urban Streetlight Image Collection for
                   Visual Monitoring and Drift Analysis (Part 2/2)
                  },
  month        = dec,
  year         = 2025,
  publisher    = {Zenodo},
  doi          = {10.5281/zenodo.17859120},
  url          = {https://doi.org/10.5281/zenodo.17859120},
}

\end{document}